\documentclass[runningheads]{llncs}
\usepackage[T1]{fontenc}
\usepackage{graphicx}
\usepackage[figurename=Fig.,labelfont=bf,labelsep=period,justification=raggedright]{caption}
\usepackage{subcaption}
\usepackage{amsmath}
\usepackage{amssymb}
\usepackage{algorithm}
\usepackage{algpseudocode}
\usepackage{float}
\usepackage{tikz}
\usepackage{float}
\usepackage[hidelinks]{hyperref}
\usetikzlibrary{shapes.geometric, arrows, calc, intersections}

\tikzstyle{startstop} = [rectangle, rounded corners, 
minimum width=3cm, 
minimum height=1cm,
text centered, 
draw=black, 
fill=red!30]

\tikzstyle{io} = [trapezium, 
trapezium stretches=true, 
trapezium left angle=70, 
trapezium right angle=110, 
minimum width=3cm, 
minimum height=1cm, text centered, 
draw=black, fill=blue!30]

\tikzstyle{process} = [rectangle, 
minimum width=3cm, 
minimum height=1cm, 
text centered, 
text width=3cm, 
draw=black, 
fill=orange!30]

\tikzstyle{decision} = [diamond, 
minimum width=2cm, 
minimum height=1cm, 
text centered, 
draw=black, 
fill=green!30]
\tikzstyle{arrow} = [thick,->,>=stealth]
\begin{document}
\title{CoreDeep: Improving Crack Detection Algorithms Using Width Stochasticity}
%
%
\author{Ramkrishna Pandey\orcidID{0000-0002-5791-6007} \and
Akshit Achara\orcidID{0009-0006-4724-5621}}
\authorrunning{R. Pandey et al.}
%
\institute{GE Research, Bangalore, India}
\maketitle              
\begin{abstract}
Automatically detecting or segmenting cracks in images can help in reducing the cost of maintenance or operations. Detecting, measuring and quantifying cracks for distress analysis in challenging background scenarios is a difficult task as there is no clear boundary that separates cracks from the background. Some of the other perceptually noted challenges with the images are variations in color, intensity, depth, blur, motion-blur, orientation, region of interest (ROI) for the defect, scale, illumination, backgrounds, etc. These variations occur across (crack inter-class) and within images (crack intra-class variabilities). In this work, we have proposed a stochastic width (SW) approach to reduce the effect of these variations. Our proposed approach improves detectability and significantly reduces false positives and negatives. We have measured the performance of our algorithm objectively in terms of mean intersection over union (mIoU) and subjectively in terms of perceptual quality.

\end{abstract}
\section{Introduction}	
Cracks are a common defect category in many distress analysis tasks ranging from road inspection to estimating the remaining useful life of the material, object, surface or machine parts. They develop in structures that are subjected to cyclic load and fatigue stress over time. Crack detection in civil engineering structures is an essential aspect of ensuring the safety and longevity of infrastructure.

There are various methods for crack detection, including visual inspection, acoustic emission testing, ultrasonic testing, infrared thermography, and ground penetrating radar. However, the proposed work is focused on visual inspection.

Visual inspection is the most commonly used method for crack detection, where trained personnel visually inspect the structure for visible cracks, and the severity of the crack is evaluated based on the thickness, length, and location of the crack.

In most of the cases, experts are needed to perform the visual inspection of cracks. This manual process is very tedious, challenging and error prone. Early detection of cracks allows us to take preventive measures for possible failures~\cite{earlydetection}. Numerous computer vision methods have been applied to detect cracks in an image. These algorithms can be broadly categorized into thresholding based ~\cite{thresholdingbasedcrackdetection}, handcrafted feature based such as local binary pattern~\cite{localbinarypattern}, histogram of oriented gradients~\cite{hogbased}, wavelets~\cite{waveletbased} and gabor filters~\cite{gaborfilter}. These approaches capture the local information but miss the global context. Crack detection using global view ~\cite{globalview} takes in account geometric and photo-metric characteristics that help in better connectivity. However, the performance drops significantly in challenging backgrounds. 

Recently, deep learning techniques have shown promising results in various computer vision tasks such as semantic segmentation, object detection and classification. These algorithms extract features in hierarchical form from a low level to a high level representation. These features are extracted from data to capture local as well as global feature representation of an image that helps in solving many challenging problems. In~\cite{tenconpandeyer}, the authors have shown that the performance of algorithms can be improved at three different levels i.e. input, architecture, and the objective followed by post-processing. In~\cite{bdisr}, the authors have shown randomly dropping character pixels in input image can help better connecting strokes thereby improving OCR performance. In this work, we have explored input with better augmentation strategy~\cite{randaug}, used two different backbones ~\cite{resnet50}, ~\cite{efficienet} as the encoder of U-Net~\cite{unet}. We have used binary cross entropy~\cite{bce} and binary focal dice loss~\cite{seg models} (objective exploration), to improve the performance of our crack detection algorithm by a significant margin.  
\subsection{Contributions}
Following are our key contributions :  

1. We advocate the use of stochastic dilation (also referred as stochastic width, see Figure~\ref{fig:stochasticwidth}) to improve the crack detectability and connectivity (see results shown in Figure~\ref{fig:resnetUnet model results} and~\ref{fig:efficientnetUnet model results}).\\
2. We use binary focal dice loss~\cite{seg models} and random augmentation~\cite{randaug} to improve the performance over our own baseline by a significant margin. \\
3. We have analyzed the prediction probabilities and grouped the predictions with different probability ranges to obtain a threshold cutoff probability (see subsection \ref{threshold selection}) that further refines the predictions and improves the mIoU.\\
4. Our method significantly reduces false positives (FP) and false negatives (FN) ( refer section ~\ref{false_p_n} for details on the computation ) by a significant margin (see Tables~\ref{table:overallresults_resnet50} and~\ref{table:overallresults_efficientnetb4}). 

\section{Datasets used for study}\label{datasets}

We have used Kaggle crack segmentation dataset ~\cite{kagglecracksegmentationdataset}, ~\cite{zhang2016road}, ~\cite{yang2019feature}, ~\cite{eisenbach2017how}, ~\cite{shi2016}, ~\cite{Amhaz2016}, ~\cite{globalview}  that has around 11298 images. These images are obtained after merging 12 crack segmentation datasets. including some 'noncrack' images (images that have no cracks). All the images are resized to a fixed size of (448, 448). Dataset is split into a train and test folders. The train folder contains 9603 images and test folder contains 1695 images. The splitting is stratified so that the proportion of each datasets in the train and test folder are similar.

\subsection{Training and Validation}
\label{train_val}

For all our experiments, we have further divided the train split of the dataset into train and validation splits. We created the validation set from the overall train set by aggregating 20 percent of the images from each of the 12 crack segmentation datasets. The validation set remains fixed for all the experiments.

The model trained on overall dataset using the binary cross-entropy loss function~\cite{bce} is hereafter referred as 'Baseline' and model trained on overall dataset with random augmentations during the training process using binary focal dice loss function~\cite{seg models} is hereafter referred as 'Baseline$+$'.

\subsection{Stochastic Width (SW)}
\label{swsection}

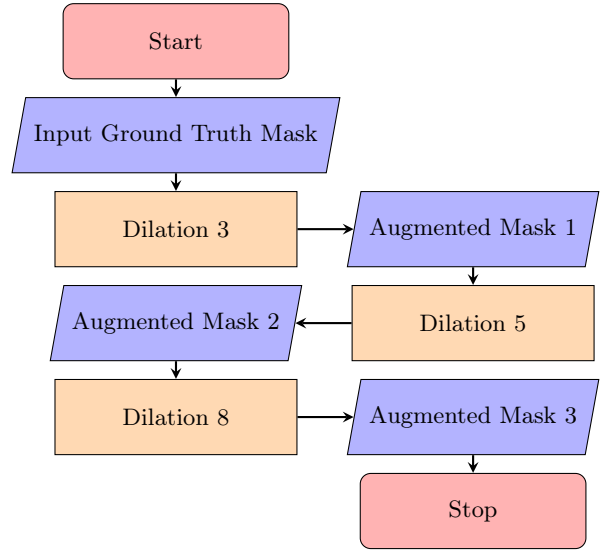
\begin{figure}[t]
	\centering
	\begin{tikzpicture}[node distance=1.25cm, scale=0.5]
		
		\node (start) [startstop] {Start};
		\node (in1) [io, below of=start] {Input Ground Truth Mask};
		\node (pro1) [process, below of=in1] {Dilation 3};
		\node (out1) [io, right of=pro1, xshift=2.7cm] {Augmented Mask 1};
		\node (pro2) [process, below of=out1] {Dilation 5};
		\node (out2) [io, left of=pro2, xshift=-2.7cm] {Augmented Mask 2};
		\node (pro3) [process, below of=out2] {Dilation 8};	
		\node (out3) [io, right of=pro3, xshift=2.7cm] {Augmented Mask 3};
		\node (stop) [startstop, below of=out3] {Stop};
		
		\draw [arrow] (start) -- (in1);
		\draw [arrow] (in1) -- (pro1);
		\draw [arrow] (pro1) -- (out1);
		\draw [arrow] (out1) -- (pro2);
		\draw [arrow] (pro2) -- (out2);
		\draw [arrow] (out2) -- (pro3);
		\draw [arrow] (pro3) -- (out3);	
		\draw [arrow] (out3) -- (stop);
		
	\end{tikzpicture}
	\caption{The flowchart shows the process stochastic width augmentation. The input ground truth mask is augmented multiple times using Dilation 3, Dilation 5 and Dilation 8 incrementally. The augmented dataset will include 4 masks (Input Ground Truth Mask, Augmented Mask 1, Augmented Mask 2 and Augmented Mask 3) for each input.} \label{fig:SWAUG}
\end{figure}

\begin{figure}[b]
	\centering
	\includegraphics[width=0.95\textwidth,height=0.15\textheight]{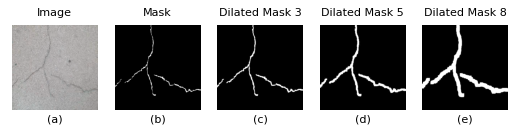}
	\caption{Shows (a) original image, (b) ground truth, (c) dilated with mask 3$\times$3, (d) dilated (c) mask 5$\times$5 and (e) dilated (d) mask 8$\times$8.}
	\label{fig:stochasticwidth}
\end{figure}

We perform augmentation on the training data by taking each input ground truth mask and performing augmentations of $3\times3$, $5\times5$ and $8\times8$ incrementally as shown in figure~\ref{fig:SWAUG}. Figure~\ref{fig:stochasticwidth} shows the different masks obtained for an input image in the stochastic width approach.

We trained a model on the above mentioned augmented dataset which is hereafter referred as 'SW'. The loss function used here is binary focal dice loss~\cite{seg models}.

We assume that (a) there is fuzziness in the boundaries of crack which can easily result in human error while marking the crack pixels, and (b) there is a variability in the width of the crack which is seen across different datasets (e.g. see image (i) in figures~\ref{fig:cfd} and~\ref{fig:crack500} taken from two different datasets. Figure~\ref{fig:cfd} has thinner cracks as compared to ~\ref{fig:crack500}) and within each image (e.g  top right portion of the crack the image (a) in figure~\ref{fig:cfd} is thin and faint compared to other cracks in the same image). Therefore, we incorporate width stochasticity to allow the model to learn the width variabilities across and within multiple datasets. 

\section{Experiments}
\label{experiments}

We have performed all the experiments using the 2D Unet architectures initialised with resnet50~\cite{resnet50} and efficientnet~\cite{efficienet} backbones trained on imagenet~\cite{imagenet} dataset.

We also perform runtime augmentations during our training process by selecting two augmentations randomly~\cite{randaug} from flip, rotate, brightness contrast, shiftScaleRotate, shear, scale, translate, multiplicativeNoise, randomGamma, downScale, HueSaturationValue, CLAHE, channel dropout, coarse dropout~\cite{coarsedropout}, color jitter, gaussian blur, median blur, grid dropout and maskDropout and, applying them incrementally(one after the other) on each input. We have used a batch size of 8 for training all our models. The optimizer used is adam~\cite{adam} with an initial learning rate of 1e-3. We used the ReduceLROnPlateau~\cite{keraslib} learning rate callback to adjust the learning rate based on the change in validation loss. After every 50 epochs, the learning rate was reduced by 0.5 with a minimum possible learning rate of 1e-6.

\subsection{Threshold Selection}
\label{threshold selection}

We have used the numpy~\cite{numpy} implementation of argmax function for calculating the labels from prediction probabilities. We have also analyzed the prediction probabilities from the model to obtain the predictions with high probabilities (strong candidate pixels for cracks) and the predictions with low probability (weak candidate pixels for cracks). We have performed an experiment on the validation dataset where we obtain the crack probabilities for all the pixels in an image. For each validation image prediction, the pixel crack probabilties are divided into ten equally sized bins which are created using the hist function from matplotlib~\cite{matplotlib}. The bins are arranged in ascending order of probabilities. We start from minimum probability value and keep increasing the threshold (to the maximum probability of the next bin) till the number of connected components (cracks) is more than the number of connected components in the initial prediction. Once the number of connected components in the thresholded prediction are more than the original prediction, the current bin is further divided into 10 bins and the threshold is reduced from the maximum probability of the last bin to the first bin until the number of connected components is same as in the original prediction.

\begin{figure}[t]
	\centering
	\includegraphics[width=0.8\textwidth,height=0.5\textheight]{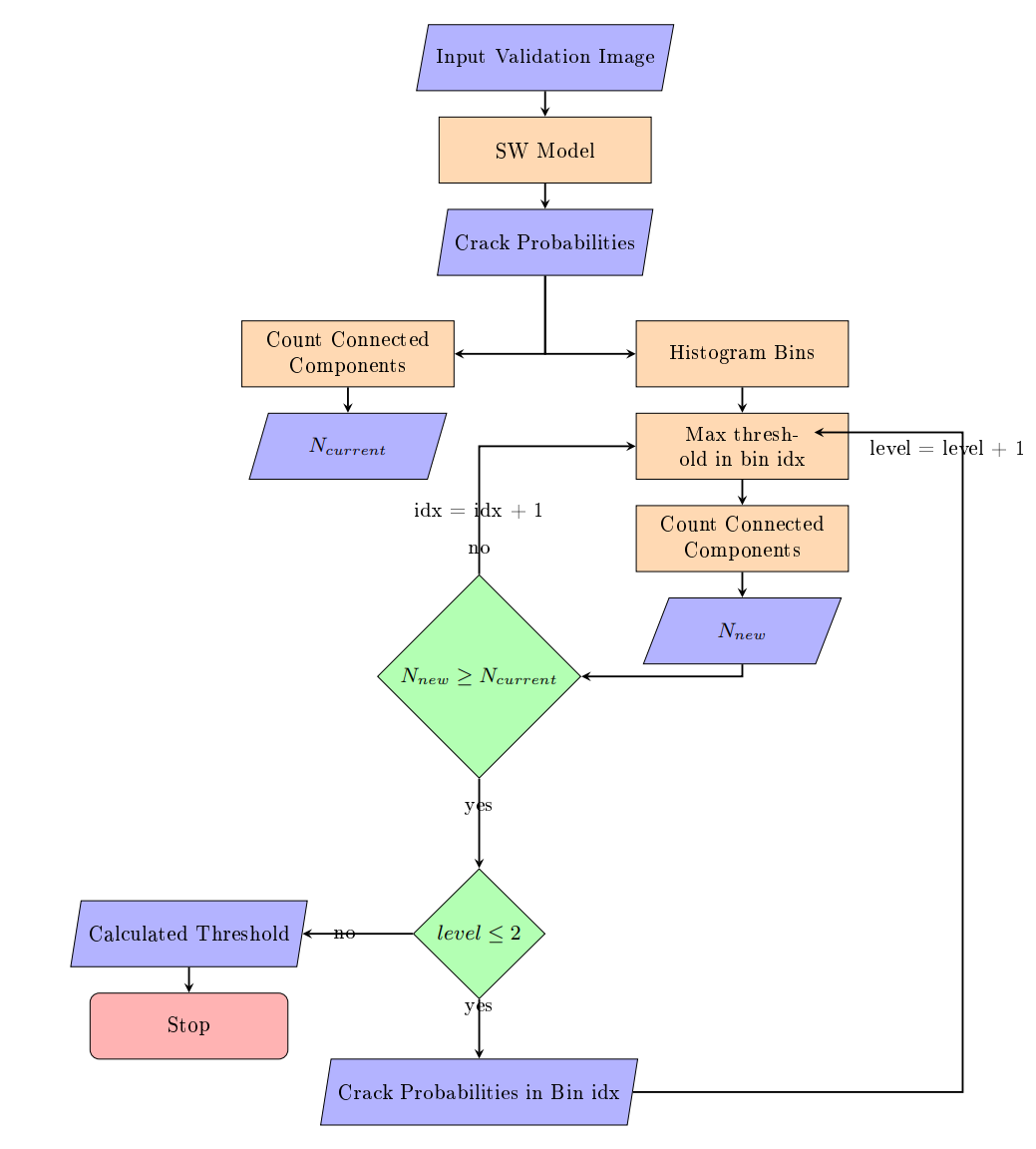}
	\caption{The flowchart shows the process of calculating the threshold probability for each validation image as discussed in section~\ref{threshold selection}. $idx$ is the bin index ($0$ to $9$) and $level$ is the number of times (maximum of 2 times) binning is done. $N\_current$ is the number of connected components in the original prediction and $N\_new$ is the number of connected components after applying a threshold on the original prediction. The calculated thresholds are analyzed to get the final threshold to use on test data (see section~\ref{threshold selection}).} \label{fig:THCAL}
\end{figure}

Finally, we created 10 bins from all the thresholds and found that the most common threshold values lie in the 10th bin, so we chose the middle value of 0.95 from the last bin boundaries (0.899 and 0.999). Figure~\ref{fig:THCAL} shows a flowchart for calculating the thresholds for each validation image. The overall idea here is to remove the pixels along the crack boundaries and avoid disconnectivity of the cracks.

The threshold values of 0.90 to 0.98 with a step size of 0.01 were used on the test image predictions and it was observed that the weaker predictions were removed resulting in a significant increase in the mean IoU as shown in the tables~\ref{table:overallresults_resnet50}, ~\ref{table:individualresults_resnet50},~\ref{table:overallresults_efficientnetb4} and~\ref{table:individualresults_efficientnetb4}. The thresold value of 0.95 resulted in the highest increase in the IoU.

\subsection{Naming Conventions}
\label{ncon}

The SW outputs post thresholding will hereafter be referred as $SW$ $<$T$>$ where T is the $threshold$ $applied$ $\times 100$. For Example, $SW$ $95$ means the output is obtained using the model trained on the augmented dataset on which a threshold of 0.95 is applied.

\subsection{Image Level False Positives and False Negatives}
\label{false_p_n}
The approach to compute false positives and false negatives listed in the tables~\ref{table:overallresults_resnet50} and~\ref{table:overallresults_efficientnetb4} are as follows; if the model predicts a crack even if there was no crack in the true mask and the image is FP. So, we count the number of 'noncrack' datasets images for which there was a crack predicted. The false negatives are computed by counting non crack predictions (misses) when there was actually a crack in the true mask and the image. FP and FN computation can be considered an alternative metric to access the performance of the algorithms in practical scenarios.              

\section{Results and Discussions}
\label{resultsanddiscussions}
\begin{figure*}[htpb]
	\begin{minipage}[h]{0.45\textwidth}
		\begin{subfigure}{\textwidth}
			\centering
			\includegraphics[width=\textwidth,height=0.055\textheight]{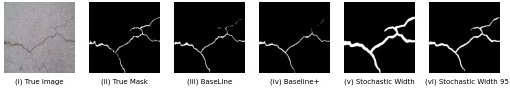}
			\caption{CFD~\cite{shi2016}.}
			\label{fig:cfd}
		\end{subfigure}
		
		\begin{subfigure}{\textwidth}
			\centering
			\includegraphics[width=\textwidth,height=0.055\textheight]{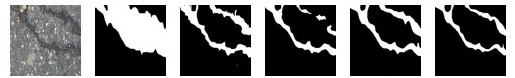}
			\caption{crack500~\cite{zhang2016road}.}
			\label{fig:crack500}
		\end{subfigure}
		
		\begin{subfigure}{\textwidth}
			\centering
			\includegraphics[width=\textwidth,height=0.055\textheight]{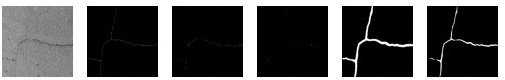}
			\caption{cracktree200.}
			\label{fig:cracktree200}
		\end{subfigure}
		
		\begin{subfigure}{\textwidth}
			\centering
			\includegraphics[width=\textwidth,height=0.055\textheight]{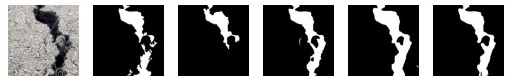}
			\caption{DeepCrack. }
			\label{fig:DeepCrack}
		\end{subfigure}
		
		\begin{subfigure}{\textwidth}
			\centering
			\includegraphics[width=\textwidth,height=0.055\textheight]{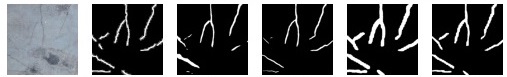}
			\caption{Eugen Muller.}
			\label{fig:Eugen_Muller}
		\end{subfigure}
		
		\begin{subfigure}{\textwidth}
			\centering
			\includegraphics[width=\textwidth,height=0.055\textheight]{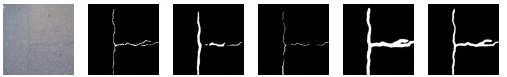}
			\caption{forest.}
			\label{fig:forest}
		\end{subfigure}
		
		\begin{subfigure}{\textwidth}
			\centering
			\includegraphics[width=\textwidth,height=0.055\textheight]{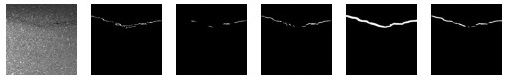}
			\caption{GAPS384~\cite{eisenbach2017how}.}
			\label{fig:GAPS}
		\end{subfigure}
		
		\begin{subfigure}{\textwidth}
			\centering
			\includegraphics[width=\textwidth,height=0.055\textheight]{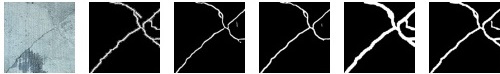}
			\caption{Rissbilder.}
			\label{fig:Rissbilder}
		\end{subfigure}
		
		\begin{subfigure}{\textwidth}
			\centering
			\includegraphics[width=\textwidth,height=0.055\textheight]{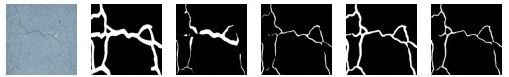}
			\caption{Sylvie Chambon.}
			\label{fig:sylvie}
		\end{subfigure}
		
		\begin{subfigure}{\textwidth}
			\centering
			\includegraphics[width=\textwidth,height=0.055\textheight]{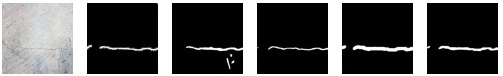}
			\caption{Volker.}
			\label{fig:volker}
		\end{subfigure}
		\captionsetup{labelformat=empty}
		\caption{All approaches here are trained using resnet50 backbone and test split of each dataset was used for predictions.}
		\label{fig:resnetUnet model results}
	\end{minipage}%
	\hfill
	\begin{minipage}[h]{0.45\linewidth}		
		\begin{subfigure}{\textwidth}
			\centering
			\includegraphics[width=\textwidth,height=0.055\textheight]{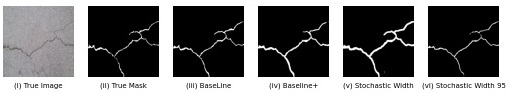}
			\caption{CFD \protect\cite{shi2016}.}
			\label{fig:cfd_e}
		\end{subfigure}
		
		\begin{subfigure}{\textwidth}
			\centering
			\includegraphics[width=\textwidth,height=0.055\textheight]{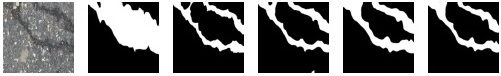}
			\caption{crack500 \protect\cite{zhang2016road}.}
			\label{fig:crack500_e}
		\end{subfigure}
		
		\begin{subfigure}{\textwidth}
			\centering
			\includegraphics[width=\textwidth,height=0.055\textheight]{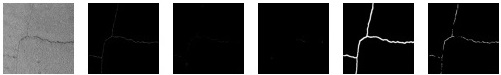}
			\caption{cracktree200 \protect\cite{globalview}.}
			\label{fig:cracktree200_e}
		\end{subfigure}
		
		\begin{subfigure}{\textwidth}
			\centering
			\includegraphics[width=\textwidth,height=0.055\textheight]{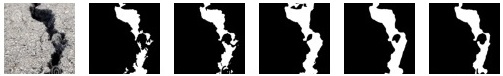}
			\caption{DeepCrack.}
			\label{fig:DeepCrack_e}
		\end{subfigure}
		
		\begin{subfigure}{\textwidth}
			\centering
			\includegraphics[width=\textwidth,height=0.055\textheight]{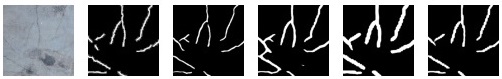}
			\caption{Eugen Muller.}
			\label{fig:Eugen_Muller_e}
		\end{subfigure}
		
		\begin{subfigure}{\textwidth}
			\centering
			\includegraphics[width=\textwidth,height=0.055\textheight]{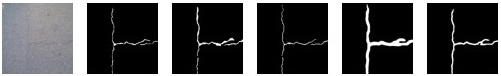}
			\caption{forest.}
			\label{fig:forest_e}
		\end{subfigure}
		
		\begin{subfigure}{\textwidth}
			\centering
			\includegraphics[width=\textwidth,height=0.055\textheight]{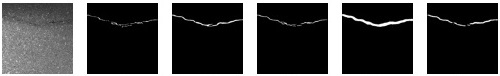}
			\caption{GAPS384 \protect\cite{eisenbach2017how}.}
			\label{fig:GAPS_e}
		\end{subfigure}
		
		\begin{subfigure}{\textwidth}
			\centering
			\includegraphics[width=\textwidth,height=0.055\textheight]{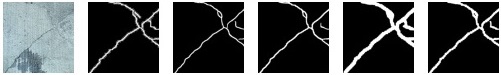}
			\caption{Rissbilder.}
			\label{fig:Rissbilder_e}
		\end{subfigure}
		
		\begin{subfigure}{\textwidth}
			\centering
			\includegraphics[width=\textwidth,height=0.055\textheight]{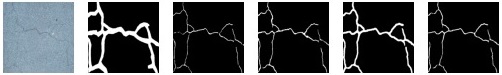}
			\caption{Sylvie Chambon~\cite{Amhaz2016}.}
			\label{fig:sylvie_e}
		\end{subfigure}
		
		\begin{subfigure}{\textwidth}
			\centering
			\includegraphics[width=\textwidth,height=0.055\textheight]{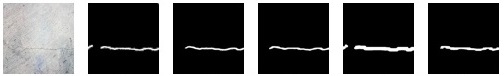}
			\caption{Volker.}
			\label{fig:volker_e}
		\end{subfigure}
		\captionsetup{labelformat=empty}
		\caption{All approaches here are trained using efficientnetb4 backbone and test split of each dataset was used for predictions.}
		\label{fig:efficientnetUnet model results}
	\end{minipage}
	\caption{Figures show an example image from each dataset; from left to right : (a) original image, (b) ground truth, (c) Baseline, (d) Baseline+ and (e) SW (f) SW 95 (see section~\ref{ncon} for details on naming). Zoom to see details.}
	\label{fig:model_results}
\end{figure*}

The results reported in the tables~\ref{table:overallresults_resnet50},~\ref{table:individualresults_resnet50},~\ref{table:overallresults_efficientnetb4} and~\ref{table:individualresults_efficientnetb4} and figures~\ref{fig:resnetUnet model results} and~\ref{fig:efficientnetUnet model results} are obtained on the test split of the kaggle crack segmentation dataset~\cite{kagglecracksegmentationdataset} (for more details, see section~\ref{datasets}).  

Table~\ref{table:overallresults_resnet50} shows the quantitative results obtained by the models trained with different approaches utilizing resnet50~\cite{resnet50} backbone. The SW approach shows a significant improvement in terms of all the metrics (except FN, which increases slightly) on applying thresholding (see the approaches SW 94, SW 95 and SW 96). The false positives have significantly reduced from 32 to 2 for SW 95 approach. 

Table~\ref{table:overallresults_efficientnetb4} shows the quantitative results obtained by the models trained with different approaches utilizing efficientnetb4~\cite{efficienet} backbone. It can be observed that these approaches have significantly less false positives as compared to the approaches in table~\ref{table:individualresults_resnet50} which utilize resnet50~\cite{resnet50} backbone. Thresholding on SW results in an improvement in all metrics except for false negatives which is similar to the pattern in table~\ref{table:overallresults_resnet50}. 

Note that a false positive here is calculated with respect to an image i.e. a prediction is called as false positive if it containts crack pixels and the ground truth has no crack pixles (there are 'noncrack' images in the dataset which do not have any cracks). Similarly, FN (false negative) in tables~\ref{table:individualresults_resnet50} and~\ref{table:individualresults_efficientnetb4} 
is calculated with respect to an image i.e. a prediction is called as false negative if it contains no crack pixels and the ground truth contains crack pixels. 

Overall, the SW 95 approach has the highest crack IoU and crack F1 amongst all the approaches.

Tables~\ref{table:individualresults_resnet50} and~\ref{table:individualresults_efficientnetb4} show the crack IoU of our approaches over multiple datasets. It can be observed that thresholding results in a higher crack IoU as compared to SW over all datasets except Sylive Chambon where it reduces.

\begin{table}[t]
	\tiny
	\caption{The tables show a comparison of our proposed approaches using multiple metrics. mIoU is mean IoU, mF1 is mean F1 score, FN is the count of false negative images, FP is the count of false positive images, C\_IoU is crack IoU, C\_P is crack precision, C\_R is crack recall, C\_F1  is crack F1 score and B\_F1 is background F1 score.}
	\label{table:overall_results}
	\begin{subtable}{0.95\linewidth}
		\caption{The results here are obtained using the 2D Unet models with an resnet50 \protect\cite{resnet50} backbone as the encoder.}
		\label{table:overallresults_resnet50}
		\resizebox{\linewidth}{!}{%
			\begin{tabular}{|lllllllllll|}
				\hline
				Approach &  mIoU &   mF1 &  FN &  FP &  C\_IoU &  B\_IoU &  C\_P &  C\_R &  C\_F1 &  B\_F1 \\
				\hline
				Baseline &  0.73 &  0.82 &  20 &  32 &      0.485 &            0.98 &             0.73 &          0.60 &     0.649 &           0.99 \\
				Baseline+ &  0.75 &  0.83 &  22 &  30 &      0.515 &            0.98 &             0.72 &          0.65 &     0.676 &           0.99 \\
				SW &  0.71 &  0.80 &  16 &  11 &      0.464 &            0.96 &             0.51 &          0.89 &     0.624 &           0.98 \\
				SW 94 &  0.75 &  0.83 &  21 &   2 &      0.526 &            0.98 &             0.67 &          0.76 &     0.680 &           0.99 \\
				\textbf{SW 95} &  0.75 &  0.83 &  22 &   2 &      0.526 &            0.98 &             0.68 &          0.74 &     0.680 &           0.99 \\
				SW 96 &  0.75 &  0.83 &  22 &   1 &      0.525 &            0.98 &             0.70 &          0.71 &     0.678 &           0.99 \\
				\hline
				
			\end{tabular}
		}%
	\end{subtable}
	\begin{subtable}{0.95\linewidth}
		\caption{The results here are obtained using the 2D Unet models with an efficientnetb4 \protect\cite{efficienet} backbone as the encoder.}
		\label{table:overallresults_efficientnetb4}
		\resizebox{\linewidth}{!}{%
			\begin{tabular}{|lllllllllll|}	
				\hline
				Approach &  mIoU &   mF1 &  FN &  FP &  C\_IoU &  B\_IoU &  C\_P &  C\_R &   C\_F1 &  B\_F1 \\
				\hline		
				BaseLine &  0.76 &  0.84 &   5 &   2 &  0.543 &   0.98 &         0.78 &      0.65 &  0.689 &  0.99 \\
				BaseLine+ &  0.77 &  0.85 &  15 &   6 &  0.557 &   0.98 &         0.73 &      0.71 &  0.705 &  0.99 \\
				SW &  0.71 &  0.80 &   8 &   4 &  0.458 &   0.96 &         0.49 &      0.93 &  0.615 &  0.98 \\
				SW 94 &  0.76 &  0.84 &  11 &   2 &  0.550 &   0.98 &         0.66 &      0.81 &  0.699 &  0.99 \\
				\textbf{SW 95} &  0.76 &  0.84 &  12 &   1 &  0.553 &   0.98 &         0.68 &      0.78 &  0.701 &  0.99 \\
				SW 96 &  0.77 &  0.84 &  11 &   1 &  0.552 &   0.98 &         0.70 &      0.75 &  0.700 &  0.99 \\
				\hline
			\end{tabular}
		}%
	\end{subtable}
\end{table}
\vfill
\begin{table}[t]
	\tiny
	\caption{The tables show a comparison of the crack IoU of all the methods on various datasets.}   
	\label{table:individualresults}
	\begin{subtable}{0.98\linewidth}
		\caption{The results here are obtained on multiple test datasets using the 2D Unet models with n resnet50 \protect\cite{resnet50} backbone as the encoder.}
		\label{table:individualresults_resnet50}
		\resizebox{\linewidth}{!}{%
			\begin{tabular}{|lllllll|}
				\hline
				Dataset      &  BaseLine &  BaseLine+ &     SW &  SW 94 &  \textbf{SW 95} &  SW 96 \\
				\hline
				Sylvie Chambon       &     0.262 &      0.165 &  0.495 &  0.298 &  0.278 &  0.254 \\
				Eugen Muller        &     0.342 &      0.548 &  0.531 &  0.626 &  0.625 &  0.619 \\
				CRACK500     &     0.552 &      0.584 &  0.578 &  0.555 &  0.547 &  0.534 \\
				cracktree200 &     0.109 &       0.04 &  0.059 &  0.113 &  0.121 &  0.131 \\
				Volker       &     0.585 &      0.601 &  0.548 &   0.66 &  0.664 &  0.666 \\
				DeepCrack    &     0.669 &      0.684 &  0.491 &  0.625 &  0.634 &  0.645 \\
				forest       &     0.253 &      0.467 &  0.212 &  0.322 &  0.334 &  0.349 \\
				Rissbilder   &     0.474 &      0.503 &  0.395 &  0.525 &  0.533 &   0.54 \\
				GAPS384      &     0.282 &      0.331 &  0.366 &  0.333 &  0.315 &   0.29 \\
				CFD          &     0.276 &      0.459 &  0.186 &  0.287 &  0.299 &  0.315 \\
				\hline
			\end{tabular}
		}%
	\end{subtable}
	\begin{subtable}{0.98\linewidth}
		\caption{The results are obtained on multiple test datasets using the 2D Unet models with an efficientnetb4 \protect\cite{efficienet} backbone as the encoder. }
		\label{table:individualresults_efficientnetb4}
		\resizebox{\linewidth}{!}{%
			\begin{tabular}{|lllllll|}
				\hline
				Dataset      &  BaseLine &  BaseLine+ &     SW &  SW 94 &  \textbf{SW 95} &  SW 96 \\
				\hline
				Sylvie Chambon      &     0.229 &      0.401 &  0.452 &  0.277 &  0.256 &   0.24 \\
				Eugen Muller       &     0.602 &      0.612 &  0.496 &  0.618 &  0.625 &  0.631 \\
				CRACK500     &     0.591 &      0.625 &  0.574 &  0.609 &  0.604 &  0.596 \\
				cracktree200 &     0.076 &      0.028 &  0.079 &  0.176 &  0.179 &  0.169 \\
				Volker       &      0.65 &       0.66 &  0.527 &  0.656 &  0.663 &   0.67 \\
				DeepCrack    &     0.747 &      0.726 &  0.511 &  0.685 &  0.696 &  0.705 \\
				forest       &     0.432 &      0.317 &  0.231 &  0.381 &  0.398 &  0.415 \\
				Rissbilder   &      0.51 &      0.524 &  0.381 &   0.52 &  0.529 &  0.536 \\
				GAPS384      &      0.41 &      0.413 &  0.365 &  0.374 &  0.356 &  0.327 \\
				CFD          &     0.401 &      0.301 &  0.212 &  0.356 &  0.373 &   0.39 \\
				\hline
			\end{tabular}
		}%
	\end{subtable}
\end{table} 
\subsection{Observations}\label{observations}

The images in figures~\ref{fig:resnetUnet model results} and~\ref{fig:efficientnetUnet model results} compare the Baseline, Baseline$+$, SW and SW $<$T$>$ methods. 
Figures~\ref{fig:crack500},~\ref{fig:crack500_e}, ~\ref{fig:cracktree200} and~\ref{fig:cracktree200_e} show that our predictions are better than that of the ground truth mask. Figure~\ref{fig:volker_e} shows that the left side portion of the crack is captured better in stochastic width experiment compared to the other methods.

It can be observed in the figures~\ref{fig:resnetUnet model results} and~\ref{fig:efficientnetUnet model results} that the boundary pixels in the thicker crack predictions from SW have been improved in SW 95 which is more precise. However, it is to be noted that there are some cases of disconnectivity in cracks in SW 95 predictions. The SW method captures the crack randomness effectively resulting in the least misses, thereby obtaining better connected cracks than other methods.

\begin{figure}[H]
	\begin{subfigure}{0.48\textwidth}
		\centering
		\includegraphics[width=\textwidth,height=0.12\textheight]{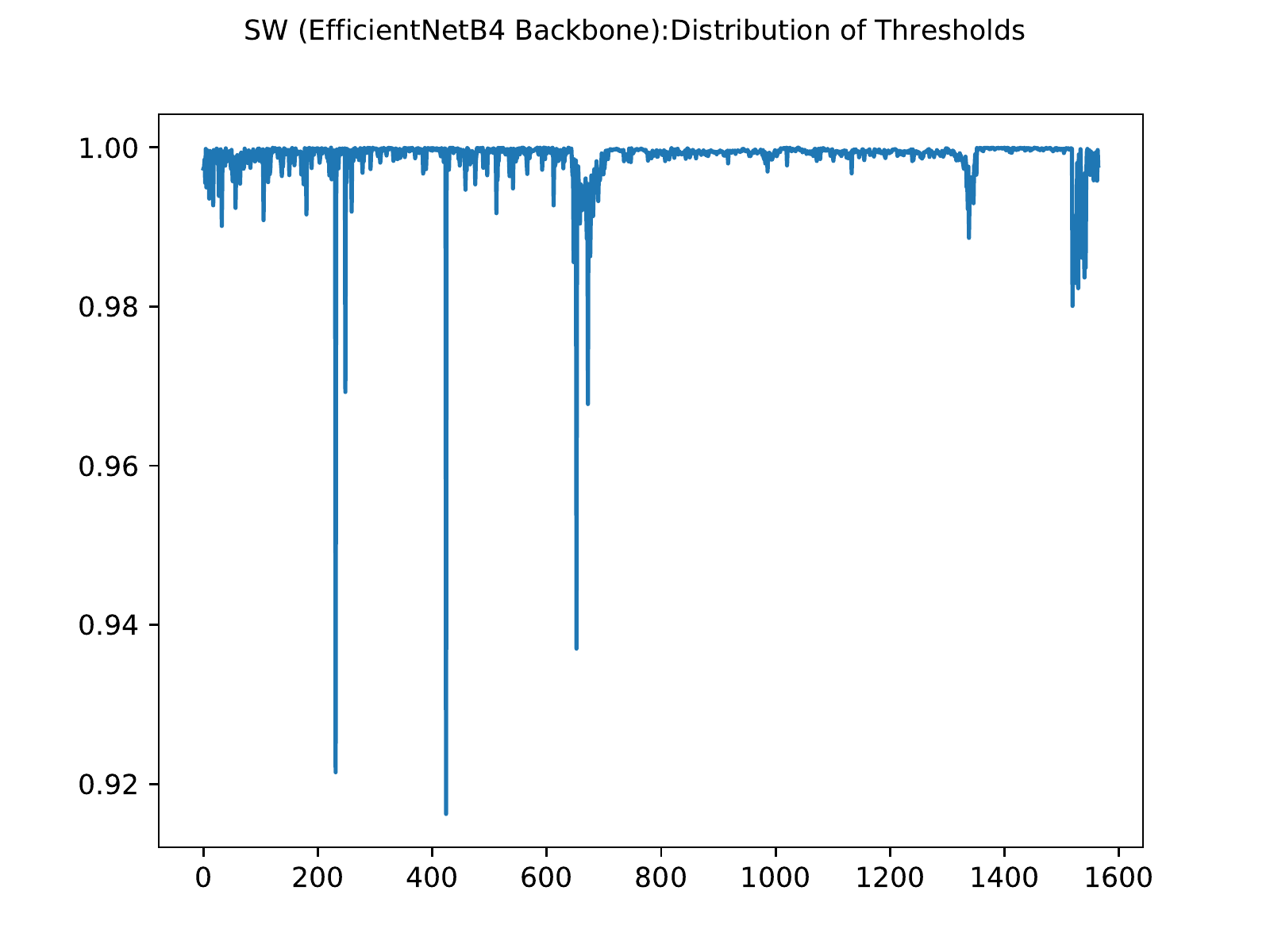}
		\caption{Shows that the threshold distribution (from the bin: 0.899 to 0.999) of pixels in the prediction mask using the SW model trained using efficientnetb4 backbone  }
		\label{fig:threshold_plot_effnet}
	\end{subfigure}
	\begin{subfigure}{0.48\textwidth}
		\centering
		\includegraphics[width=\textwidth,height=0.12\textheight]{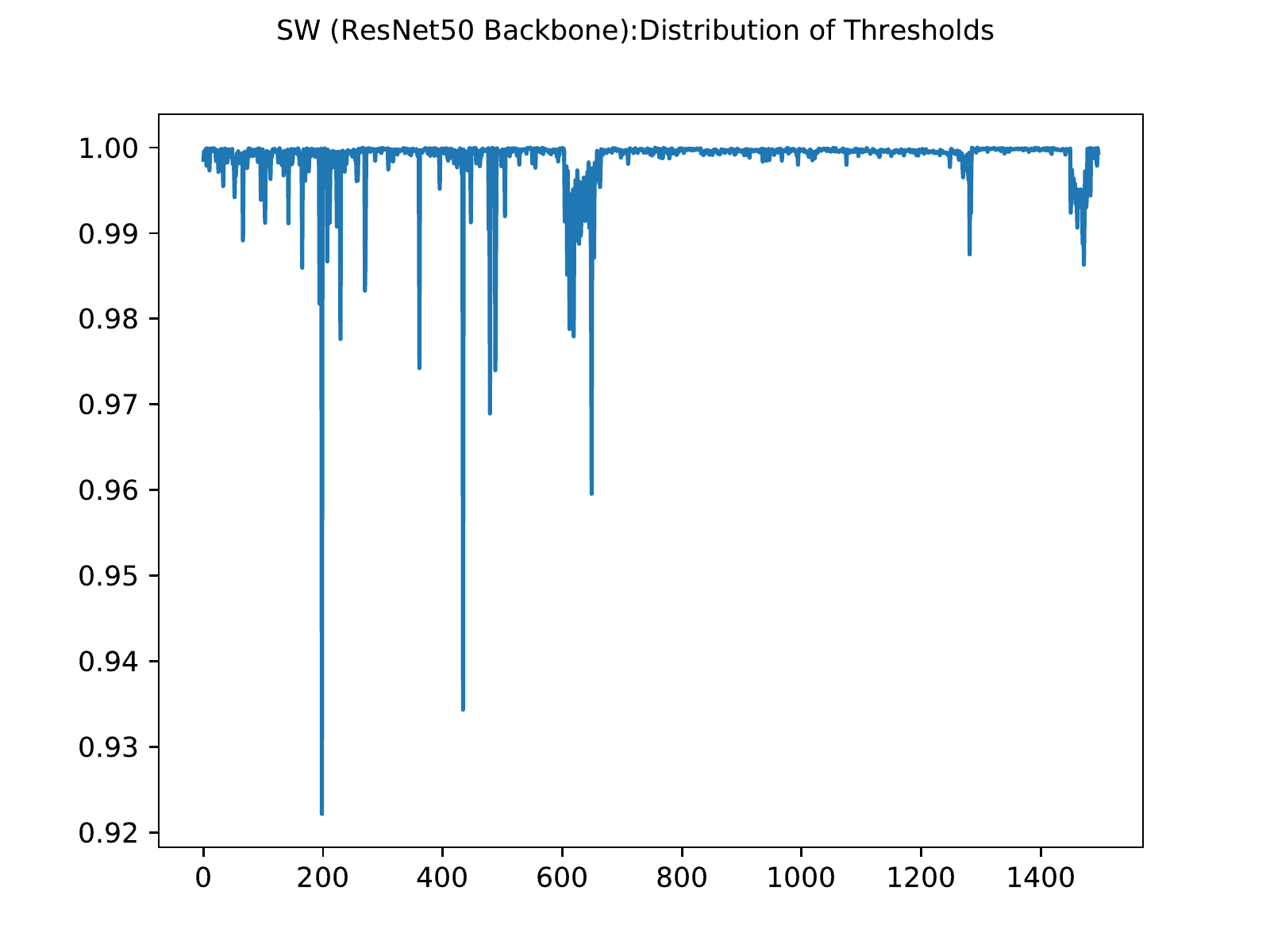}
		\caption{Shows that the threshold distribution (from all bins: 0.899 to 0.999) of pixels in the prediction mask using the SW model trained using resnet50 backbone }
		\label{fig:threshold_plot_resnet}
	\end{subfigure}
	\begin{subfigure}{0.48\textwidth}
		\centering
		\includegraphics[width=\textwidth,height=0.12\textheight]{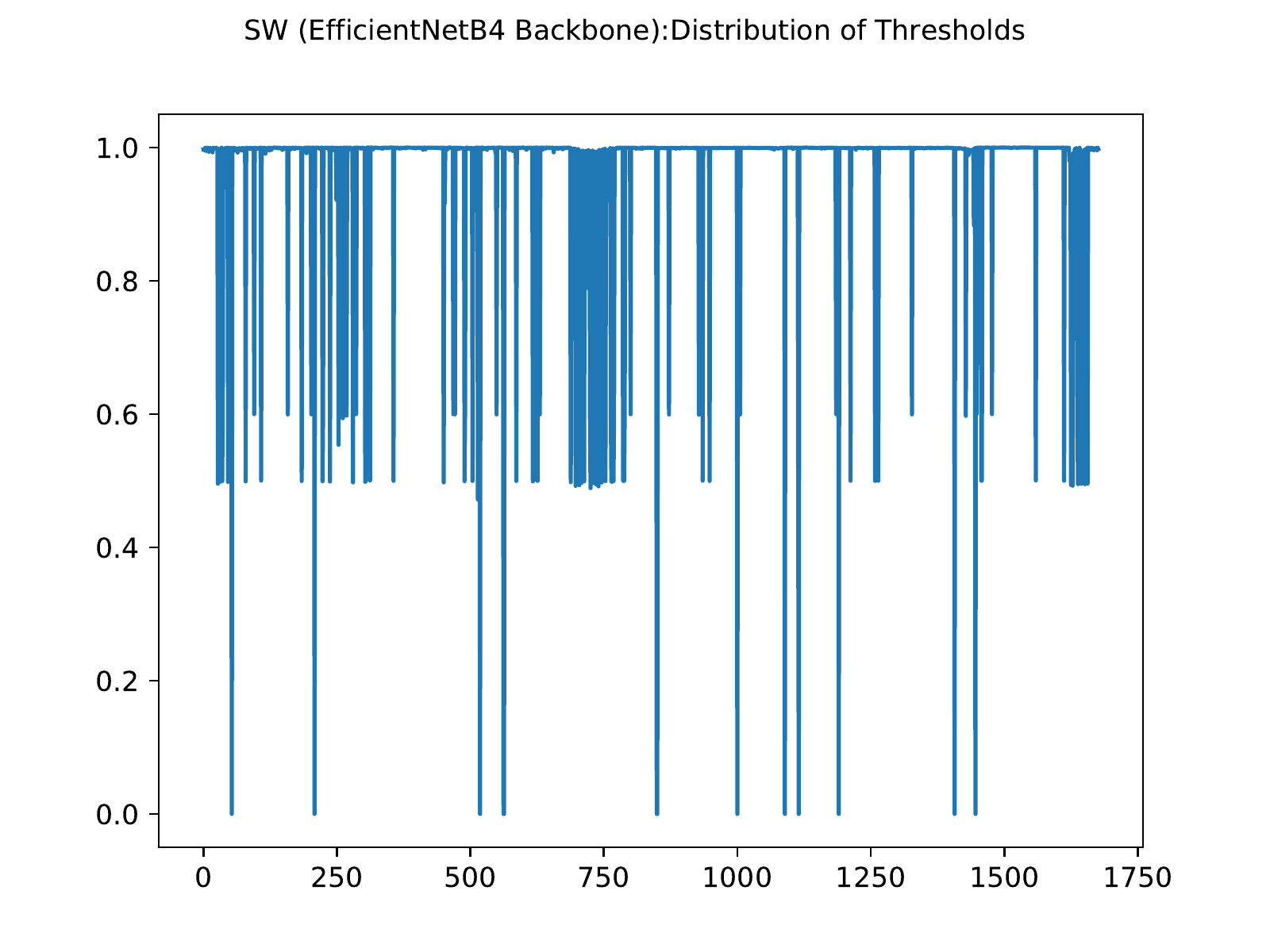}
		\caption{Shows that the threshold distribution (from the bin: 0 to 0.999) of pixels in the prediction mask using the SW model trained using efficientnetb4 backbone  }
		\label{fig:threshold_plot_effnet_entire}
	\end{subfigure}
	\hfill
	\begin{subfigure}{0.48\textwidth}
		\centering
		\includegraphics[width=\textwidth,height=0.12\textheight]{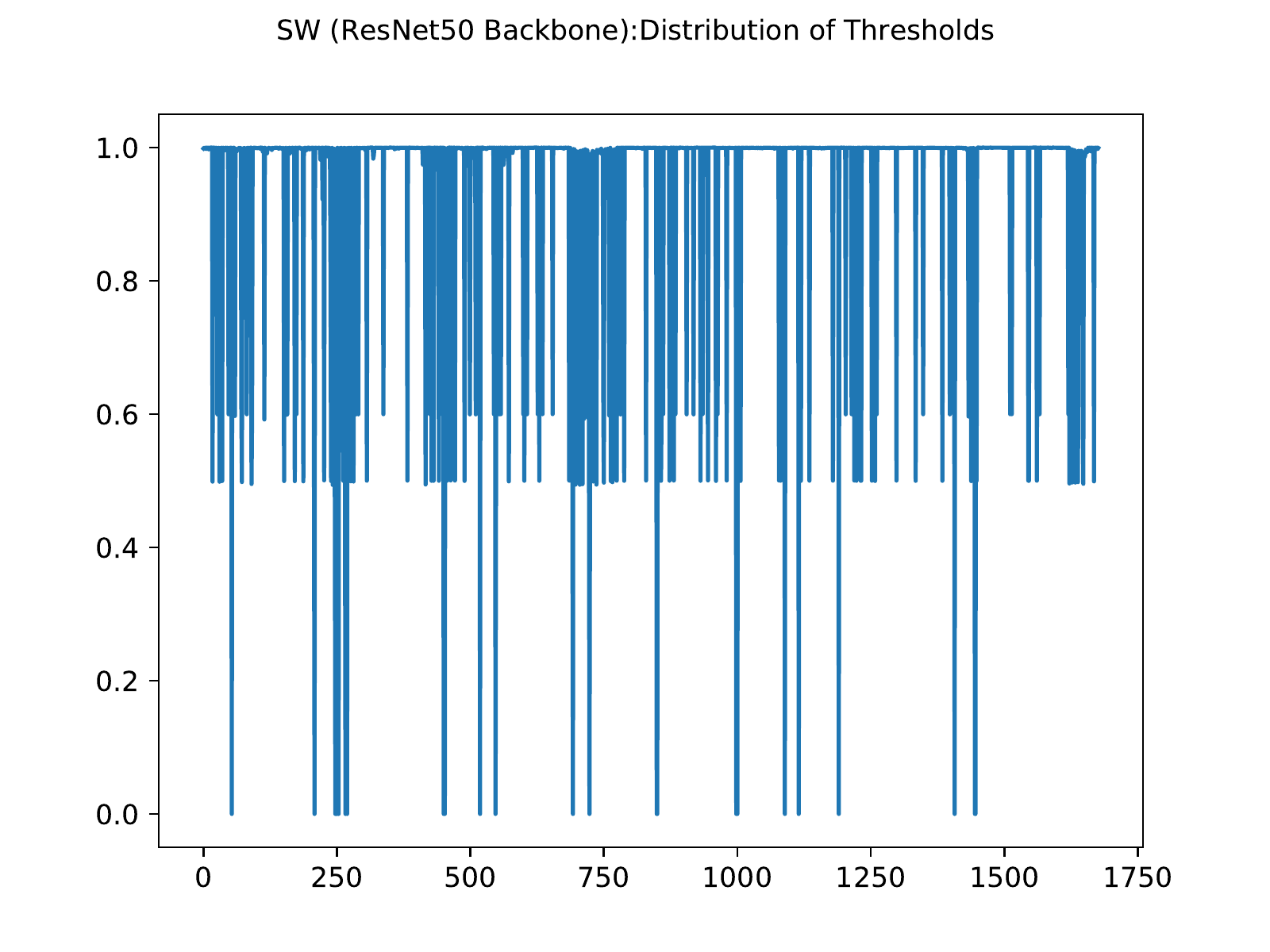}
		\caption{Shows that the threshold distribution (from all bins: 0 to 0.999) of pixels in the prediction mask using the SW model trained using resnet50 backbone }
		\label{fig:threshold_plot_resnet_entire}
	\end{subfigure}
	\caption{Shows the plot of threshold for each image in the val set (crack images)}
	\label{fig:thresholdallplots}
\end{figure}

Figures~\ref{fig:threshold_plot_effnet} and~\ref{fig:threshold_plot_resnet}  show the frequency plot all the crack images in the validation dataset having thresholds above 0.899 using the models trained with efficientnetb4~\cite{efficienet} and resnet50~\cite{resnet50} backbones respectively. Figures~\ref{fig:threshold_plot_effnet_entire} and~\ref{fig:threshold_plot_resnet_entire} show the frequency plot all the crack images in the validation dataset over all thresholds using the models trained with efficientnetb4~\cite{efficienet} and resnet50~\cite{resnet50} backbones respectively.

It can be observed that most of the crack probabilities are above 0.899 and efficientnetb4 model has around 1600 images while resnet50 model has around 1450 images above threshold 0.899.  This implies that efficient net model is better compared to the resnet50 model. The plot also visually validates the idea of binning discussed in section~\ref{threshold selection}.

\section{Conclusion}
We have shown that judiciously combining traditional approaches in a deep learning framework can result in significant boost in the performance of the crack detection algorithms. The proposed SW method results in better crack connectivity, reduces false positives and improves the crack detactability by a significant margin (see results in the figures~\ref{fig:resnetUnet model results} and~\ref{fig:efficientnetUnet model results} and tables~\ref{table:overallresults_resnet50},~\ref{table:overallresults_efficientnetb4}). We have further refined the predictions based on the probabilities of each pixel in predicted masks. The thresholded masks obtain better estimates of "crack width" which further refines the predictions, improves the perceptual quality (shown in figures~\ref{fig:resnetUnet model results} and~\ref{fig:efficientnetUnet model results}), reduces the number of false positives (in terms of pixels), and improves the mIoU by a significant margin as listed in the tables~\ref{table:overallresults_resnet50} and ~\ref{table:individualresults_efficientnetb4}.

\subsubsection{Acknowledgements}
We would like to thank GE Research for supporting our work. We would also like to thank the publishers of the kaggle crack segmentation dataset~\cite{kagglecracksegmentationdataset} which enabled us to conduct this study.


\begin{thebibliography}{00}
	
	\bibitem{earlydetection}
	Dhital, D., and Lee, J. R. A fully non-contact ultrasonic propagation imaging system for closed surface crack evaluation. Experimental mechanics, 52(8), 1111-1122 (2012).
	
	\bibitem{thresholdingbasedcrackdetection}
	Liu, Fanfan, Guoai Xu, Yixian Yang, Xinxin Niu, and Yuli Pan. "Novel approach to pavement cracking automatic detection based on segment extending." In 2008 International Symposium on Knowledge Acquisition and Modeling, pp. 610-614. IEEE, 2008.	
	
	\bibitem{localbinarypattern}
	Hu, Yong, and Chun-xia Zhao. "A novel LBP based methods for pavement crack detection." Journal of pattern Recognition research 5, no. 1 (2010): 140-147.
	
	\bibitem{hogbased}
	Kapela, Rafał, Paweł Śniatała, Adam Turkot, Andrzej Rybarczyk, Andrzej Pożarycki, Paweł Rydzewski, Michał Wyczałek, and Adam Błoch. "Asphalt surfaced pavement cracks detection based on histograms of oriented gradients." In 2015 22nd International Conference Mixed Design of Integrated Circuits \& Systems (MIXDES), pp. 579-584. IEEE, 2015.	
	
	\bibitem{waveletbased}
	Zhou, Jian, Peisen S. Huang, and Fu-Pen Chiang. "Wavelet-based pavement distress detection and evaluation." Optical Engineering 45, no. 2 (2006): 027007-027007.
	
	\bibitem{gaborfilter}
	Xu, Hongyan, Xiu Su, Yi Wang, Huaiyu Cai, Kerang Cui, and Xiaodong Chen. "Automatic bridge crack detection using a convolutional neural network." Applied Sciences 9, no. 14 (2019): 2867.
	
	\bibitem{globalview}
	Zou, Qin, Yu Cao, Qingquan Li, Qingzhou Mao, and Song Wang. "CrackTree: Automatic crack detection from pavement images." Pattern Recognition Letters 33, no. 3 (2012): 227-238.
	
	\bibitem{tenconpandeyer}
	Pandey, Ram Krishna, A. G. Ramakrishnan, and Souvik Karmakar. "Effects of modifying the input features and the loss function on improving emotion classification." In TENCON 2019-2019 IEEE Region 10 Conference (TENCON), pp. 1159-1162. IEEE, 2019.
	
	\bibitem{bdisr}
	Pandey, Ram Krishna, K. Vignesh, and A. G. Ramakrishnan. "Binary document image super resolution for improved readability and OCR performance." arXiv preprint arXiv:1812.02475 (2018).
	
	\bibitem{randaug}
	Cubuk, Ekin D., Barret Zoph, Jonathon Shlens, and Quoc V. Le. "Randaugment: Practical automated data augmentation with a reduced search space." In Proceedings of the IEEE/CVF conference on computer vision and pattern recognition workshops, pp. 702-703. 2020.
	
	\bibitem{resnet50}
	He, Kaiming, Xiangyu Zhang, Shaoqing Ren, and Jian Sun. "Deep residual learning for image recognition." In Proceedings of the IEEE conference on computer vision and pattern recognition, pp. 770-778. 2016.
	
	\bibitem{efficienet}
	Tan, Mingxing, and Quoc Le. "Efficientnet: Rethinking model scaling for convolutional neural networks." In International conference on machine learning, pp. 6105-6114. PMLR, 2019.
	
	\bibitem{unet}
	Ronneberger, Olaf, Philipp Fischer, and Thomas Brox. "U-net: Convolutional networks for biomedical image segmentation." In Medical Image Computing and Computer-Assisted Intervention–MICCAI 2015: 18th International Conference, Munich, Germany, October 5-9, 2015, Proceedings, Part III 18, pp. 234-241. Springer International Publishing, 2015.
	
	\bibitem{bce}
	Zhang, Zhilu, and Mert Sabuncu. "Generalized cross entropy loss for training deep neural networks with noisy labels." Advances in neural information processing systems 31 (2018).
	
	\bibitem{seg models}
	Iakubovskii, Pavel. Segmentation Models. GitHub. GitHub repository. 
	https://github.com/qubvel/segmentation\_models (2019).
	
	\bibitem{kagglecracksegmentationdataset}
	'Kaggle Crack Segmentation Dataset', https://www.kaggle.com/datasets/lakshaymiddha/crack-segmentation-dataset, accessed 2020
	
	\bibitem{zhang2016road} 
	Zhang, Lei, Fan Yang, Yimin Daniel Zhang, and Ying Julie Zhu. "Road crack detection using deep convolutional neural network." In 2016 IEEE international conference on image processing (ICIP), pp. 3708-3712. IEEE, 2016.
	
	\bibitem{yang2019feature}
	Yang, Fan, Lei Zhang, Sijia Yu, Danil Prokhorov, Xue Mei, and Haibin Ling. "Feature pyramid and hierarchical boosting network for pavement crack detection." IEEE Transactions on Intelligent Transportation Systems 21, no. 4 (2019): 1525-1535.
	
	\bibitem{eisenbach2017how}
	Eisenbach, Markus, Ronny Stricker, Daniel Seichter, Karl Amende, Klaus Debes, Maximilian Sesselmann, Dirk Ebersbach, Ulrike Stoeckert, and Horst-Michael Gross. "How to get pavement distress detection ready for deep learning? A systematic approach." In 2017 international joint conference on neural networks (IJCNN), pp. 2039-2047. IEEE, 2017.
	
	\bibitem{shi2016}
	Shi, Yong, Limeng Cui, Zhiquan Qi, Fan Meng, and Zhensong Chen. "Automatic road crack detection using random structured forests." IEEE Transactions on Intelligent Transportation Systems 17, no. 12 (2016): 3434-3445.
	
	\bibitem{Amhaz2016}
	Amhaz, Rabih, Sylvie Chambon, Jérôme Idier, and Vincent Baltazart. "Automatic crack detection on two-dimensional pavement images: An algorithm based on minimal path selection." IEEE Transactions on Intelligent Transportation Systems 17, no. 10 (2016): 2718-2729.
	
	\bibitem{imagenet}
	Deng, Jia, Wei Dong, Richard Socher, Li-Jia Li, Kai Li, and Li Fei-Fei. "Imagenet: A large-scale hierarchical image database." In 2009 IEEE conference on computer vision and pattern recognition, pp. 248-255. Ieee, 2009.
	
	\bibitem{coarsedropout}
	DeVries, Terrance, and Graham W. Taylor. "Improved regularization of convolutional neural networks with cutout." arXiv preprint arXiv:1708.04552 (2017).
	
	\bibitem{adam}
	Kingma, Diederik P., and Jimmy Ba. "Adam: A method for stochastic optimization." arXiv preprint arXiv:1412.6980 (2014).
	
	\bibitem{keraslib}
	Chollet, F., \& others.  Keras. GitHub. 
	Retrieved from https://github.com/fchollet/keras (2015).
	
	\bibitem{numpy}
	Harris, Charles R., K. Jarrod Millman, Stéfan J. Van Der Walt, Ralf Gommers, Pauli Virtanen, David Cournapeau, Eric Wieser et al. "Array programming with NumPy." Nature 585, no. 7825 (2020): 357-362.
	
	\bibitem{matplotlib}
	Hunter, John D. "Matplotlib: A 2D graphics environment." Computing in science \& engineering 9, no. 03 (2007): 90-95.
	
	\bibitem{albumentations}
	Buslaev, Alexander, Vladimir I. Iglovikov, Eugene Khvedchenya, Alex Parinov, Mikhail Druzhinin, and Alexandr A. Kalinin. "Albumentations: fast and flexible image augmentations." Information 11, no. 2 (2020): 125.

\end{thebibliography}
\end{document}